\documentclass{article}
\usepackage{spconf,amsmath,graphicx}
\usepackage[fontsize=9pt]{scrextend}

\usepackage{booktabs}
\usepackage{balance}

\usepackage{amsmath,epsfig,amssymb,amsthm,url,amsfonts}
\usepackage{algorithm}
\usepackage{algpseudocode}
\usepackage{multirow}
\usepackage{mdframed}
\usepackage{url}
\usepackage{enumitem}
\usepackage[makeroom]{cancel}
\usepackage{dblfloatfix}
\usepackage{array}
\usepackage{comment}
\usepackage{epstopdf}
\usepackage{float}
\usepackage{subfig}
\usepackage{breqn}
\usepackage{cite}
\usepackage{xcolor}
\usepackage{cases}
\usepackage{tabularx}
\usepackage[printonlyused,withpage]{acronym}
\usepackage{balance}
\usepackage{graphbox} 

\usepackage{hyperref}
\setlist[itemize]{label=$\triangleright$}
\newtheoremstyle{break}
{}
{}
{\itshape}
{}
{\bfseries}
{.}
{\newline}
{}
\theoremstyle{break}

\theoremstyle{definition}

\newcommand{\vect}[1]{\mathbf{#1}}
\newcommand{\bs}[1]{\boldsymbol{#1}}
\newcommand{\E}{\mathbb{E}}
\usepackage{url}

\makeatletter
\def\thmhead@plain#1#2#3{%
	\thmname{#1}\thmnumber{\@ifnotempty{#1}{ }\@upn{#2}}%
	\thmnote{ {\the\thm@notefont#3}}}
\let\thmhead\thmhead@plain
\makeatother

\graphicspath{{./figs/}}

\newcommand{\argmax}{\operatornamewithlimits{argmax}}
\newcommand{\argmin}{\operatornamewithlimits{argmin}}

\newcommand{\lk}{ \left\{ }
\newcommand{\rk}{ \right\} }

\newcommand{\diag}{\mbox{{diag}}}

\newsavebox\mybox

\acrodef{SE}{speech enhancement}
\acrodef{STFT}{short-time Fourier transform}
\acrodef{ESTOI}{extended short-time objective intelligibility}
\acrodef{NMF}{non-negative matrix factorization}
\acrodef{DNN}{deep neural network}
\acrodef{VAE}{variational auto-encoder}
\acrodefplural{VAEs}{variational auto-encoders}
\acrodef{EM}{expectation-maximisation}
\acrodef{TF}{time-frequency}
\acrodef{ELBO}{evidence lower bound}
\acrodef{LR}{Living Room}
\acrodef{SDR}{signal-to-distortion ratio}
\acrodef{PESQ}{perceptual evaluation of speech quality}
\acrodef{SNR}{signal-to-noise ratio}
\acrodef{DNNs}{deep neural networks}
\acrodef{VESDE}{variance-preserving stochastic differential equation}
\acrodef{SDE}{stochastic differential equation}
\acrodef{GAN}{generative adversarial networks}
\acrodefplural{GANs}{generative adversarial networks}
\acrodef{SI-SDR}{scale-invariant signal-to-distortion ratio}
\acrodef{MOS}{mean opinion score}
\acrodef{SGMSE+}{score-based generative model for speech enhancement}
\acrodef{NCSNPP++}{Noise-Conditional Score Network}
\acrodef{WSJ}{Wall Street Journal}
\acrodef{UDiffSE}{Unsupervised Diffusion-Based Speech Enhancement}
\acrodef{PC}{Predictor-Corrector}
\acrodef{DMPS}{Diffusion Model Posterior Sampling}
\acrodef{NN}{neural network}

\newcommand{\score}[1]{\vect{S}_{\theta^*}(\vect{s}_{#1}, {#1})}

\newcommand{\normal}{\mathcal{N}(\vect{0}, \vect{I})}
\newcommand{\normalc}{\mathcal{N}_{\mathbb{C}}(\vect{0}, \vect{I})}

\usepackage{bm}

\title{Unsupervised Speech Enhancement with Diffusion-based Generative Models}
\name{%
Berné Nortier, %
Mostafa Sadeghi,
Romain Serizel %
\thanks{This work was supported by the French National Research Agency (ANR) under the project REAVISE (ANR-22-CE23-0026-01). Experiments presented in this paper were carried out using the Grid'5000 testbed, supported by a scientific interest group hosted by Inria, and including CNRS, RENATER, and several universities as well as other organizations (see https://www.grid5000.fr). }}
\address{%
Université de Lorraine, CNRS, Inria, LORIA, F-54000 Nancy, France}
\begin{document}
%
\maketitle
\begin{abstract}
Recently, conditional score-based diffusion models have gained significant attention in the field of supervised speech enhancement, yielding state-of-the-art performance. However, these methods may face challenges when generalising to unseen conditions. To address this issue, we introduce an alternative approach that operates in an unsupervised manner, leveraging the generative power of diffusion models. Specifically, in a training phase, a clean speech prior distribution is learnt in the \ac{STFT} domain using score-based diffusion models, allowing it to unconditionally generate clean speech from Gaussian noise. Then, we develop a posterior sampling methodology for speech enhancement by combining the learnt clean speech prior with a noise model for speech signal inference. The noise parameters are simultaneously learnt along with clean speech estimation through an iterative \ac{EM} approach. To the best of our knowledge, this is the first work exploring diffusion-based generative models for unsupervised speech enhancement, demonstrating promising results compared to a recent \ac{VAE}-based unsupervised approach and a state-of-the-art diffusion-based supervised method. It thus opens a new direction for future research in unsupervised speech enhancement.
\end{abstract}
\section{Introduction}
\label{sec:intro}
Over the past decade, the \ac{SE} task has been extensively investigated, and numerous novel approaches have been proposed that greatly leverage the advancements and efficacy of \ac{DNN} architectures \cite{wang2018supervised}. The majority of these approaches are based on supervised (discriminative) learning of a \ac{DNN} over training pairs of clean and noisy speech signals, covering different speakers, noise types, and \ac{SNR} values. Such an approach depends heavily on the number and diversity of training samples and noise conditions, and thus generalisation to unseen (out-of-domain) environments cannot be guaranteed. 

Unsupervised \ac{SE} based on deep generative models presents an alternative approach with improved generalisation performance~\cite{bando2018statistical, bie2022unsupervised, bando2020adaptive}. In contrast to purely supervised methods, the generative-based (unsupervised) framework learns the statistical distribution of clean speech signals and uses it as a prior distribution for inferring the target signal from its noisy observation. In these methods, \ac{VAE} \cite{KingW14} has been commonly used as a generative clean speech prior, which is combined with a \ac{NMF}-based observation model to estimate clean speech following a statistical \ac{EM} framework. 

Recently, diffusion-based generative models have emerged as a powerful and state-of-the-art framework to model complex data distributions \cite{sohl2015deep, song2021scorebased}. These models learn an implicit distribution by estimating the score, i.e., the gradient of the log probability density (with respect to data). This is done by gradually diffusing data samples into noise and then learning a score approximating model that can reverse the noising process for different noise scales. The forward process of corrupting data is modelled as a \ac{SDE}, which can be reversed and yields a corresponding reverse \ac{SDE} that depends only on the score of the perturbed data and may easily be solved numerically. Diffusion-based models have been widely applied to the \ac{SE} task in a supervised way \cite{lu2022conditional, serra2022universal, welker22speech, richter2023speech, yen2023cold} by incorporating noisy speech signals in the diffusion process as conditioning information.

In this paper, we develop an \textit{unsupervised} speech enhancement framework leveraging diffusion-based generative models as data-driven priors. Specifically, in a training step, the statistical characteristics of \textit{clean speech signals} are learnt in the complex \ac{STFT} domain through the use of a score-based diffusion model. At test time, we perform posterior sampling by combining the learnt implicit clean speech prior with a parametric statistical model for noise to infer the clean speech signal. The noise parameters are estimated alongside the clean speech signal by following an iterative \ac{EM}-based approach. To our knowledge, this is the first work that proposes using diffusion-based generative models for unsupervised \ac{SE}, and explores their potential. We conduct experiments comparing the proposed framework with a VAE-based unsupervised approach \cite{bie2022unsupervised} as well as a state-of-the-art diffusion-based supervised method \cite{richter2023speech}. The results demonstrate the effectiveness and promising performance of the proposed diffusion-based unsupervised approach, paving the path for future research in this direction.

The rest of the paper is organised as follows: Section~\ref{sec:sbdm} reviews score-based diffusion modelling and \ac{VAE}-based \ac{SE} as two closely related problems to our work. The proposed speech generative modelling and enhancement frameworks are detailed in Section~\ref{sec:prop}. Experimental results are then presented in Section~\ref{sec:exp}, followed by a conclusion and suggestions for future lines of work in Section~\ref{sec:conc}.

\vspace{-2mm}
\section{Background}
\label{sec:sbdm}

\subsection{Score-based diffusion models}
Diffusion models are a state-of-the-art class of probabilistic generative models that have recently achieved remarkable performance in generating high-quality samples in different applications \cite{song2021scorebased}. These models transform an unknown data distribution $p_{0}$ to a tractable prior distribution, usually $\normal$, by gradually adding noise to training data in a forward process. Then, in a reverse process, a parameterised model is learnt to iteratively generate samples starting from noise and transform these into samples from the unknown data distribution. This action of smoothly injecting noise into training samples may be described by a \ac{SDE}. 
Specifically, consider a diffusion process $\{\vect{s}_{t}\}_{t\in[0,1]}$, indexed by a continuous time-step variable $t$, which solves the following general linear \ac{SDE}
\begin{equation}\label{eqn:sde-fwd}
    \textrm{d}\vect{s}_t = \vect{f}(\vect{s}_t) \textrm{d}t + g(t) \textrm{d}\vect{w},
\end{equation}
where $\vect{w}$ denotes a standard Wiener process, the vector-valued $\vect{f}$ is the \textit{drift} coefficient term, and the scalar function $g$ is the \textit{diffusion} coefficient. Here, the forward process transforms a clean training sample $\vect{s}_{0}=\vect{s}$ to a noise sample $\vect{s}_{1}$, whose distribution converges to $p_1 \sim \normal$. Under some light regularity conditions~\cite{anderson1982}, the SDE in \eqref{eqn:sde-fwd} also has an associated \textit{reverse}-time \ac{SDE}: 
\begin{equation}\label{eqn:rev-sde}
    \textrm{d}\vect{s}_t = [\vect{f}(\vect{s}_t) \textrm{d}t - g(t)^2 \nabla_{\vect{s}_t} \log p_t(\vect{s}_t) ] \textrm{d}t + g(t) \textrm{d}\vect{\bar{w}},
\end{equation}
where $\vect{\bar{w}}$ is a standard Wiener process running backwards in time, $\textrm{d}t$ is an infinitesimal negative time-step, and $\nabla_{\vect{s}_t} \log p_t(\vect{s}_t)$ is called the \textit{score} function. In practice, the score is approximated by a time-dependent \ac{NN} $\score{t} \approx \nabla_{\vect{s}_t}\log p_t(\vect{s}_t)$, called the \textit{score model}, where $\theta^*$ denotes the learnt weights of the \ac{NN}. By plugging the score model in \eqref{eqn:rev-sde}, we can solve the resulting \ac{SDE} using a variety of solvers to sample from the unknown data distribution \cite{song2021scorebased}. In this paper, we make use of the \ac{PC} sampler \cite{song2021scorebased}.

\subsection{VAE-based unsupervised speech enhancement}
Previous work on unsupervised \ac{SE} use \ac{VAE} to learn the prior distribution of clean speech signals, which is then combined with an observation model to estimate clean speech in a statistical framework. Specifically, in the \ac{STFT} domain, a latent variable-based generative model is assumed as $p_{\theta}(\vect{s}, \vect{z})=p_{\theta}(\vect{s}| \vect{z})p_{\theta}(\vect{z})$, where $\vect{s}$ denotes the \ac{STFT} representation of clean speech and $\vect{z}$ represents the associated (latent) low-dimensional embedding. Some parameterised Gaussian forms for the generative distributions are usually assumed, whose parameters are learnt over clean speech data, following the evidence lower-bound optimisation principle \cite{KingW14}.

For \ac{SE}, it is assumed that $\vect{x} = \vect{s} + \vect{n}$, where $\vect{x}$, $\vect{s}$, and $\vect{n}$ denote \ac{STFT} representations of noisy (mixture) speech, clean speech, and background noise, respectively. The likelihood $p_{\phi}(\vect{x}|\vect{s})$ is usually a proper complex Gaussian distribution $\mathcal{N}_{\mathbb{C}}$ with mean $\vect{s}$, whose variance is parameterised with a low-rank \ac{NMF} factorisation. \ac{SE} then amounts to inferring the latent variable $\vect{z}$ associated with $\vect{s}$ from $\vect{x}$, which necessitates learning the \ac{NMF} parameters, denoted $\phi$, via an \ac{EM} process formulated below
\begin{equation}\label{eq:em_vae}
    \max_{\phi}~\E_{p_\phi(\vect{z}|\vect{x})}\lk\log p_\phi(\vect{x}| \vect{z})\rk.
\end{equation}
This could be solved using, e.g., the variational \ac{EM} procedure developed in \cite{leglaive2020recurrent, bie2022unsupervised}, which approximates $p_\phi(\vect{z}|\vect{x})$.
\section{Proposed framework}
\label{sec:prop}
\subsection{Diffusion-based speech generative modeling}\label{sec:diff_speech}
Following~\cite{richter2023speech}, we work with the complex-valued \ac{STFT} representations of speech signals and apply an exponential amplitude transformation to balance the heavy-tailed distribution of \ac{STFT} amplitudes. 

Like VAE, the diffusion-based generative model is independently defined for each \ac{TF} bin. Therefore, as done in~\cite{richter2023speech}, all the vector-valued variables $\vect{s}_t$ in boldface contain flattened \ac{TF} representations of speech signals. For concrete instantiations of the forward and reverse \ac{SDE} (~\eqref{eqn:sde-fwd} and \eqref{eqn:rev-sde} respectively), we use an alternative form of the well-known \ac{VESDE}~\cite{song2019generative} inspired by~\cite{richter2023speech}, and adapt it to obtain the drift and diffusion coefficients as follows
 \begin{align}\label{eqn:drift-diff-ve}
    \vect{f}(\vect{s}_t)=-\gamma\vect{s}_t, 
    \hspace{7mm} 
    g(t)= \sigma_{\textrm{max}} \Bigl(\frac{\sigma_{\textrm{max}}}{ \sigma_{\textrm{min}}}\Bigr)^{t}\sqrt{2 \log \Bigl(\frac{\sigma_{\textrm{max}}}{ \sigma_{\textrm{min}}}\Bigr)},
\end{align}
where $\gamma$ is a constant parameter, and $\sigma_{\textrm{min}}$ and $\sigma_{\textrm{max}}$ are parameters defining the noise schedule of the Wiener process. The \ac{SDE} in \eqref{eqn:sde-fwd} then has the \textit{perturbation kernel} defined below, which allows one to sample $\vect{s}_t$ directly given $\vect{s}_0$
\begin{equation}\label{eqn:p0t-richter}
    p_{0t}(\vect{s}_t | \vect{s}_0) = \mathcal{N}_{\mathbb{C}}({\delta}_t \vect{s}_0, \sigma(t)^2\vect{I}),
\end{equation}
where ${\delta}_t = \textrm{e}^{-\gamma t}$ and the variance term $\sigma(t)^2$ is given by
\begin{equation}\label{eqn:var-richter}
    \sigma(t)^2 = \frac{\sigma_{\textrm{min}}^{2} \Big((\sigma_{\textrm{max}}/ \sigma_{\textrm{min}})^{2t} - {\delta_t}^{2} \Big){\log(\sigma_{\textrm{max}}/ \sigma_{\textrm{min}})}}{\gamma + \log(\sigma_{\textrm{max}}/ \sigma_{\textrm{min}})}.
\end{equation}
To learn the \ac{NN} parameters $\theta$, a weighted Fisher divergence~\cite{song2019generative} between the true and approximated score is solved, which,  after some mathematical manipulation, leads to the following training objective~\cite{richter2023speech}
\begin{equation}\label{eqn:train-obj}
    \theta^{*} = {\argmin_{\theta} \mathbb{E}_{t, \vect{s}, \bs{\zeta}, \vect{s}_t | \vect{s}}}
    {\Bigr[\| {\vect{S}_{\theta}(\vect{s}_{t}, {t})}+ \frac{\bs{\zeta}}{\sigma(t)} \|_2^2 \Bigl]},
\end{equation}
where $\bs{\zeta}\sim\mathcal{N}_{\mathbb{C}}(\vect{0}, \vect{I})$, i.e., complex-valued Gaussian noise. 

\subsection{Diffusion-based unsupervised speech enhancement}\label{ssec:se-as-ip}
We now describe the unsupervised \ac{SE} framework based on diffusion-based generative models. The prior clean speech distribution $p= p(\vect{s})$ is unknown, but can be obtained by training a diffusion-based generative model as described in Section~\ref{sec:diff_speech}, yielding an implicit prior, as opposed to the explicit VAE-based speech prior modelling framework. This implicit diffusion-based speech prior only allows for iterative sampling, without an explicit density form. As such, the \ac{SE} procedure adopted in VAE-based modelling cannot directly be used for diffusion-based learnt speech priors. Assuming the same observation model as before, i.e., $\vect{x} = \vect{s} + \vect{n}$, and \ac{NMF}-based likelihood parameterisation, we here propose to sample from the following intractable posterior distribution to estimate the clean speech $\vect{s}$ directly 
\begin{equation}
    p_{\phi}(\vect{s}|\vect{x}) \propto p_{\phi}(\vect{x}|\vect{s}) p_{\theta^*}(\vect{s}),
\end{equation} 
where $\theta^*$ denotes the diffusion model's pretrained, and thus fixed, parameters. We model the noise by $\vect{n} \sim \mathcal{N}_{\mathbb{C}}(\bs{0}, \diag(\text{vec}(\vect{W}\vect{H})))$ where $\vect{W}$, $\vect{H}$ are low-rank matrices with non-negative entries and rank $r$ and $\text{vec}(\vect{W}\vect{H})$ denotes the vectorised form of $\vect{W}\vect{H}$. 
The likelihood $p_{\phi}(\vect{x}|\vect{s})$ then writes as $p_{\phi}(\vect{x}|\vect{s})= \mathcal{N}_{\mathbb{C}}(\vect{s}, \diag(\vect{v}_{\phi}))$, where $\vect{v}_{\phi}=\text{vec}(\vect{W}\vect{H})$. Learning the NMF parameters, i.e., $\phi=\lk\vect{W}, \vect{H} \rk$, is done by solving
\begin{equation}\label{eq:em_diff}
    \max_{\phi}~\E_{p_\phi(\vect{s}|\vect{x})}\lk\log p_\phi(\vect{x}| \vect{s})\rk.
\end{equation}
An overview of the proposed \ac{UDiffSE} approach is provided in Algorithm~\ref{alg:1}. The following sections detail the E-step and M-step.

\begin{algorithm}[t!]
\caption{UDiffSE}\label{alg:1}
\begin{algorithmic}[1]
\State $\phi_0=\lk\vect{W}_{0},\vect{H}_{0}\rk$
\For{$k=1,\ldots, K$}
    \State $\hat{\vect{s}} \sim p_{\phi_{k-1}}(\vect{s}|\vect{x})$ \Comment{\textit{(E-Step)}}
    \State $\phi_k \gets \argmax_{\phi}\log p_\phi(\vect{x}| \hat{\vect{s}})$ \Comment{\textit{(M-Step)}} 
\EndFor
\State \Return $\hat{\vect{s}}$
\end{algorithmic}
\end{algorithm}
\subsubsection{E-Step}
Given a current estimate of $\phi$, the E-step (posterior sampling) entails the generation of speech samples from the posterior distribution $p_{\phi}(\vect{s}|\vect{x})$ to approximate the expectation in \eqref{eq:em_diff}. This is done via the construction of a stochastic process $\{\vect{s}_{t} | \vect{x}\}_{t\in[0,1]}$ by conditioning the original process $\{\vect{s}_{t}\}_{t\in[0,1]}$ on the observation $\vect{x}$ to obtain an estimate $\hat{\vect{s}}\sim p_\phi(\vect{s}|\vect{x})$. To this end, we modify the reverse \ac{SDE} \eqref{eqn:rev-sde} as follows
\begin{align}\label{eqn:rev-post-sde}
    \textrm{d}\vect{s}_t  &= \Big[\vect{f}(\vect{s}_t) \textrm{d}t - g(t)^2 {\nabla_{\vect{s}_t}\log p_t(\vect{s}_t|\vect{x})} \Big]  \textrm{d}t + g(t) \textrm{d}\vect{\bar{w}} \nonumber \\
    &= \Big[\vect{f}(\vect{s}_t) \textrm{d}t - g(t)^2 \big(\nabla_{\vect{s}_t}\log p_t(\vect{x}|\vect{s}_t) { + \nabla_{\vect{s}_t}\log p_t(\vect{s}_t)}\big)\Big]  \textrm{d}t \nonumber \\
    &\qquad\qquad + g(t) \textrm{d}\vect{\bar{w}}
\end{align}
where again the score function can be approximated by $\score{t}$. However, the conditional score function $\nabla_{\vect{s}_t}\log p_\phi(\vect{x}|\vect{s}_t)$ is, in fact, intractable to compute in closed form due to its dependence on time. That is,
\begin{equation}\label{eq:likelihood_x}
    p_\phi(\vect{x}|\vect{s}_t)=\int p_\phi(\vect{x}|\vect{s}_0)p_{t0}(\vect{s}_0|\vect{s}_t)\textrm{d}\vect{s}_0,
\end{equation}
where $p_{t0}(\vect{s}_0|\vect{s}_t)\propto p_{0t}(\vect{s}_t|\vect{s}_0)p(\vect{s}_0)$ is intractable. As an approximation, we follow~\cite{meng2022diffusion} and assume an uninformative prior $p(\vect{s}_0)$, which, along with \eqref{eqn:p0t-richter}, results in
\begin{equation}
    \tilde{p}_{t0}(\vect{s}_0|\vect{s}_t)=\mathcal{N}_{\mathbb{C}}\Bigl(\frac{\vect{s}_0}{{\delta}_t}, \frac{\sigma(t)^2}{{\delta}_t^2}\vect{I} \Bigr).
\end{equation}
Plugging this approximation in \eqref{eq:likelihood_x} gives us the following \textit{noise-perturbed pseudo-likelihood}
\begin{equation}\label{eqn:psllkd}
    \tilde{p}_\phi(\vect{x}|\vect{s}_t) \sim \mathcal{N}_{\mathbb{C}}\Big(\frac{\vect{s}_t}{{\delta}_t}, \frac{\sigma(t)^2}{{\delta}_t^2} \vect{I} + \diag(\bs{v}_{\phi})\Big).
\end{equation}
The conditional reverse process is then approximated as 
\begin{equation}\label{eqn:post-rev-sde}
    \begin{split}
        \textrm{d}\vect{s}_t  =  \Big[\vect{f}(\vect{s}_t) \textrm{d}t - g(t)^2 \score{t} \Big] \textrm{d}t + g(t) \textrm{d}\vect{\bar{w}} \\ - g(t)^2 \nabla_{\vect{s}_t} \log \tilde{p}_\phi (\vect{x}|\vect{s}_t)\textrm{d}t .
    \end{split}
\end{equation}
This is exactly the unconditional reverse process~\eqref{eqn:rev-sde} for sampling clean speech, plus an additional term which imposes data consistency. We use the change of variables formula and take the gradient to compute $\nabla_{\vect{s}_t} \textrm{log} \tilde{p} (\vect{x}|\vect{s}_t)$, the \textit{noise-perturbed pseudo-likelihood score} as 
\begin{equation}
    \nabla_{\vect{s}_t} \log \tilde{p} (\vect{x}|\vect{s}_t) = \frac{1}{{\delta}}_t \Big[\frac{\sigma(t)^2}{{\delta}_t^2} \vect{I} + \diag(\bs{v}_{\phi}) \Big]^{-1}(\frac{\vect{s}_t}{{\delta}_t} - \vect{x}).
\end{equation}
Lastly, we introduce an additional weighting parameter $\lambda$ to the pseudo-likelihood as in~\cite{meng2022diffusion} to balance the effect of the mixture signal on the estimated sample. We experimentally observed that performing the full posterior reverse step at each iteration enforces strongly the data consistency condition, causing the sample to converge to the mixture signal. To prevent this, we only perform the posterior step every $\ell$ iterations. We solve the reverse \ac{SDE} using a \ac{PC} sampler~\cite{song2021scorebased} - a numeric sampler consisting of a discretisation of~\eqref{eqn:rev-sde} - the \textit{predictor} - followed by a Langevin sampling step - the \textit{corrector} - to `correct` the marginal at time $t$. The overall E-step is summarised in Algorithm~\ref{alg:e-step}. The variable $\tau$ denotes discrete time-step in $[0,1]$. For simplicity, we employ the shorthand $\sigma_\tau$, $\vect{f}_\tau$,  $g_\tau$ for $\sigma(\tau)$, $\vect{f}(\tau)$,  $g(\tau)$, respectively. 

\subsubsection{M-Step}
Having obtained a clean speech estimate $\hat{\vect{s}}$ in the E-step, we now consider updating the noise parameters $\phi=\lk\vect{W}, \vect{H} \rk$ via \eqref{eq:em_diff}, approximating the expectation with a Monte-Carlo average using $\vect{s}\gets\hat{\vect{s}}$:
\begin{align}
    \phi &\leftarrow \argmax_{\lk\vect{W}, \vect{H}\rk\geq 0}~ \log p_\phi(\vect{x}|\hat{\vect{s}})  \nonumber\\
    &= \argmin_{\lk\vect{W}, \vect{H}\rk\geq 0}~  \frac{(\vect{x}-\hat{\vect{s}})^H(\vect{x}-\hat{\vect{s}})}{\bs{v}_{\phi}} + \log(\bs{v}_{\phi}),
\end{align}
where $(\cdot)^H$ denotes the conjugate transpose operation, and the division is done element-wise. The above problem can be solved using different algorithms, e.g., the multiplicative update rules \cite{fevotte2009nonnegative, sadeghi2020robust}.

\begin{algorithm}[t!]
\caption{Posterior sampling (E-step) of \ac{UDiffSE}}\label{alg:e-step}
\begin{algorithmic}[1]
\Require $N, \vect{x}, \ell, \lambda$
    \State ${\vect{s}}_1 \sim \mathcal{N}_{\mathbb{C}}(\vect{x}, \vect{I}), \Delta \tau \gets \frac{1}{N}$
    \For{$i=N,\ldots, 1$}
        \State $\tau\gets\frac{i}{N}$
        \State $\bs{\zeta}_c\sim\normalc$ \hspace{3mm} \Comment{\textit{(Corrector)}}
        \State ${\vect{s}}_{\tau} \gets \vect{s}_\tau + \epsilon_\tau {\vect{S}_{\theta^*}(\vect{s}_\tau, {\tau})} + \sqrt{2\epsilon_\tau}\bs{\zeta}_c$        
        \State $\bs{\zeta}_p\sim\normalc$ \hspace{3mm} \Comment{\textit{(Predictor)}}
        \State ${\vect{s}}_{\tau} \gets {\vect{s}}_{\tau} - \vect{f}_{\tau}\Delta{\tau} + g_{\tau}^2 \vect{S}_{\theta^*}({\vect{s}}_{\tau}, t) \Delta{\tau}+  g_{\tau}\sqrt{\Delta{\tau}}\bs{\zeta}_p $


        \If{$i \equiv 0 \pmod{\ell}$} \hspace{3mm} \Comment{\textit{(Posterior)}}
            \State $\nabla_{{\vect{s}}_{\tau}} \log \tilde{p} (\vect{x}|{\vect{s}}_{\tau}) \gets\dfrac{1}{{\delta}_\tau}\Big[\dfrac{{\sigma^2_\tau}}{{{\delta}^2_\tau}} \vect{I} + \diag(\bs{v}_{\phi}) \Big]^{-1}  
             (\dfrac{{\vect{s}}_{\tau} }{\delta_\tau} - \vect{x}) $
            \State $\vect{s}_{\tau} \gets {\vect{s}}_{\tau}  + {\lambda}{g_{\tau}^2} {\nabla_{\vect{s}_\tau} \log \tilde{p} (\vect{x}|\vect{s}_\tau)}$
        \EndIf
    \EndFor
    \State \Return $\hat{\vect{s}}={\vect{s}}_{0}$
\end{algorithmic}
\end{algorithm}

\begin{table*}[!t]
    \centering
\caption{Speech enhancement results under both matched and mismatched conditions. `S': supervised, `U': unsupervised. Bold and italicised indicate the best and second best performances, respectively. }
\resizebox{0.85\textwidth}{!}{
     \begin{tabular}{|l|c|c|c|c|c|c|c|}
    \hline
        {Method} &  {Type} & {SI-SDR} (dB) & {PESQ} & {ESTOI} & {SIG-MOS} & {BAK-MOS} & {OVR-MOS} \\ \hline \hline
        {Input (\textbf{WSJ0-QUT})} & {-} & -2.60~$\pm$~0.17 & 1.83~$\pm$~0.02 & 0.50~$\pm$~0.01 & 4.04~$\pm$~0.01 & 2.93~$\pm$~0.02 & 3.13~$\pm$~0.01 \\ \hline
         {RVAE \cite{leglaive2020recurrent, bie2022unsupervised}} & {U} & 4.39~$\pm$~0.21 & 2.20~$\pm$~0.02 & 0.59~$\pm$~0.01 & 3.88~$\pm$~0.02 & 3.32~$\pm$~0.02 & 3.13~$\pm$~0.02 \\ 
        UDiffSE (Ours) & {U} &  \textit{4.80~$\pm$~0.23} & \textit{2.21~$\pm$~0.02} & \textit{0.63~$\pm$~0.01} & \textit{4.33~$\pm$~0.01} & \textit{3.74~$\pm$~0.02} & \textit{3.74~$\pm$~0.02}	 \\ 
         {SGMSE+ \cite{richter2023speech}} & {S} &  \bf{9.41~$\pm$~0.18}   &  \bf{2.66~$\pm$~0.02} & \bf{0.77~$\pm$~0.01}  & \bf{4.48~$\pm$~0.01} & \bf{4.51~$\pm$~0.01} &   \bf{4.19~$\pm$~0.01}\\ \hline \hline
        
        {Input (\textbf{TCD-TIMIT})} & {-} & -8.74~$\pm$~0.29	& 1.84~$\pm$~0.02 & 0.35~$\pm$~0.01	& 3.52~$\pm$~0.02& 2.22~$\pm$~0.03 & 2.68~$\pm$~0.01 \\ \hline 
         {RVAE \cite{leglaive2020recurrent, bie2022unsupervised}}  & {U} & \bf{1.44~$\pm$~0.31} &	\textit{2.02~$\pm$~0.02}  &	0.35~$\pm$~0.01 &	3.08~$\pm$~0.03 &	\textit{3.18~$\pm$~0.02} &	2.61~$\pm$~0.02 \\ 
         UDiffSE (Ours) & {U} & \textit{0.37~$\pm$~0.25} & 2.01~$\pm$~0.02 & \bf{0.41~$\pm$~0.01} & \bf{3.91~$\pm$~0.01} & 2.88~$\pm$~0.03 & \textit{3.08~$\pm$~0.02}\\ 
         {SGMSE+ \cite{richter2023speech}}& {S} & -3.97~$\pm$~0.41 & \bf{2.04~$\pm$~0.03}	& \textit{0.38~$\pm$~0.01} & \textit{3.79~$\pm$~0.02} & \bf{3.43~$\pm$~0.02} & \bf{3.13~$\pm$~0.02}  \\ \hline 
    \end{tabular}}
     \label{tab:se_results}
\end{table*}


\section{Experiments}
\label{sec:exp}
In this section, we provide a performance evaluation of our proposed UDiffSE framework as compared against an unsupervised speech enhancement approach based on recurrent \ac{VAE} (RVAE)\footnote{\url{https://github.com/XiaoyuBIE1994/DVAE_SE/}}~\cite{leglaive2020recurrent, bie2022unsupervised}, as well as a state-of-the-art diffusion-based supervised \ac{SE} method, called \ac{SGMSE+}\footnote{\url{https://github.com/sp-uhh/sgmse}}~\cite{richter2023speech}.

\noindent\textbf{Evaluation Metrics}.
To measure the quality of the enhanced speech signals, we use standard instrumental evaluation metrics, including the \ac{SI-SDR} in dB \cite{le2019sdr}, the \ac{ESTOI} measure~\cite{jensen2016algorithm} ranging in $[0,1]$, and the \ac{PESQ} score~\cite{rix2001pesq} ranging in $[-0.5,4.5]$. We also use the DNS-MOS \cite{reddy2022dnsmos}, a non-intrusive speech quality metric, which provides scores for the speech quality (SIG), background noise quality (BAK), and overall quality (OVRL) of speech. For all the metrics, higher values indicate improved performance. 

\noindent\textbf{Datasets}. 
To learn the clean speech prior model, we train on the `si\_tr\_s` subset of the \ac{WSJ} corpus \cite{garofolo1993csr}, which amounts to roughly 25 hours of data. The \ac{STFT} is computed using a window size of 510, a hop-length of 128 ($\approx$ 75\% overlap), and a Hann window, which gives $F=256$ frequency bins. All signals have a sampling rate of 16kHz. To ensure similarity across samples of different length during training, subsamples are randomly selected from a \ac{STFT} transform so that we get $T=256$ time frames with start and end positions randomly generated during training.  

For performance evaluation, we use the WSJ0-QUT dataset created by~\cite{leglaive2020recurrent}, comprising 651 synthetic mixtures (about 1.5 hours of noisy speech data) which uses clean speech signals from the ‘si\_et\_ 05‘ subset of WSJ dataset and noise signals from the QUT-NOISE corpus~\cite{dean2015qut}. These include \textit{Café}, \textit{{Home}}, \textit{Street}, and \textit{Car} and have \ac{SNR} values of $-5$~dB, $0$~dB, and $5$~dB. We also evaluate generalisation capability of different methods in mismatched conditions by using pre-computed noisy versions of the TCD-TIMIT data presented in~\cite{abdelaziz2017ntcd}. This set contains noise types \textit{Living Room}  (from the second CHiME challenge \cite{vincent2013second}), \textit{White}, \textit{Car}, and \textit{Babble} (from the RSG-10 corpus \cite{steeneken1988description}) with SNR values of $-5$~dB, $0$~dB, and $5$~dB and. This yields 540 test speech signals (or approximately 45 minutes).

\noindent\textbf{Stochastic Differential Equation}.
The SDE in~\eqref{eqn:drift-diff-ve} has parameter values $\gamma=1.5, \sigma_{\textrm{min}}=0.05,  \sigma_{\textrm{max}}=0.5$. To avoid instabilities around 0, we adopt standard practice and set a minimum process time with $t_{\textrm{min}}=0.03$.

\noindent\textbf{Models architecture}. 
We adapt the \ac{SGMSE+} architecture developed in~\cite{welker22speech}, which is based on a multi-resolution U-Net structure, by zeroing out their $\vect{x}$ term and adapting the channels. RVAE consists of an encoder-decoder architecture composing bidirectional long short-term memory (BLSTM) networks. For both {RVAE} and {SGMSE+}, we use the pretrained models that are available in their associated public code repositories.%

\noindent\textbf{Training setup}.
We train the score model $\vect{S}_{\theta^*}$ for 220 epochs using an Adam optimiser with a learning rate of 0.0001 and a batch size of $16$. Our loss is an exponential moving average of the network's weights, initialised with a decay of 0.999. 

\noindent\textbf{EM settings}.
The reverse process in~\eqref{eqn:post-rev-sde} is solved using a \ac{PC} sampler with step size $\epsilon_\tau:= ({\sigma_\tau}/{2})^2$. The number of reverse sampling steps is set to $N=30$. The posterior update step is performed every $\ell=2$ steps, and the \ac{NMF} variances matrices have rank $r=4$. For each sample, we perform 5 \ac{EM} iterations. We observe that performing the same denoising procedure over $b$ samples in parallel and then averaging the result yields much better performance; we thus set the batch size to $b=4$. The weighting parameter $\lambda$ is set to 1.5. These parameter choices are motivated by an extensive set of experimental studies provided in the Supplementary Material.

\noindent\textbf{Results.} We report our \ac{SE} results in Table~\ref{tab:se_results}. Competing methods are evaluated in the matched and mismatched cases. Inspecting the results, we can make a number of conclusions: As may be expected, the supervised framework outperforms its unsupervised counterpart in the matched case, but at the cost of utilising labelled data. Our \ac{UDiffSE} framework outperforms the alternative unsupervised RVAE on almost all metrics under both matched and mismatched conditions. In particular, it achieves much higher ESTOI, SIG-MOS, and OVR-MOS scores than RVAE, which is more noticeable in the mismatched condition. 

Furthermore, the proposed \ac{UDiffSE} method outperforms the supervised \ac{SGMSE+} framework for both the {ESTOI} and SIG-MOS metrics in the mismatched condition, with a comparable OVR-MOS score. While all three frameworks have very similar PESQ results in the mismatched case, the unsupervised methods significantly outperform \ac{SGMSE+} in terms of SI-SDR (by more than 4 dB). The performance of \ac{UDiffSE} on the TCD-TIMIT dataset showcases its capacity to generalise to unseen data, which could possibly imply that it has learnt a good representation of general clean speech as the underlying prior. Supplementary material, including audio samples, is available online. \footnote{\url{https://github.com/joanne-b-nortier/UDiffSE}}

\section{Conclusion}\label{sec:conc}
In this paper, we introduce \ac{UDiffSE}, an unsupervised generative-based framework to solve the \ac{SE} task by learning an implicit prior distribution over clean speech data. We do this by defining a continuous diffusion process in the \ac{STFT} domain in the form of a conditional \ac{SDE}, and imposing an NMF-based parameterised additive noise model. An EM approach is developed to simultaneously generate clean speech and learn the noise parameters. An approximation of the likelihood term in the E-step then yields a tractable posterior sampling procedure. This method outperforms an unsupervised \ac{VAE}-based approach to \ac{SE} for almost all metrics in matched and mismatched test conditions, while showcasing better generalisation performance than a state-of-the-art diffusion-based supervised method. \ac{UDiffSE} does, however, have the disadvantage of being time-consuming, which originates from the complexity of the reverse diffusion process. Future works include speeding up the reverse process, utilising the recent advancements in diffusion-based image generation, and developing more efficient noise models. 



\bibliographystyle{IEEEbib-abbrev}
\bibliography{mybib}

\end{document}